\documentclass[sigconf, screen, nonacm]{acmart}
\AtBeginDocument{%
  }

\setcopyright{acmlicensed}
\copyrightyear{2026}
\acmYear{2026}
\acmDOI{XXXXXXX.XXXXXXX}
\acmConference[MM '26]{The 34th ACM International Conference on Multimedia}{November 10--14, 2026}{Rio de Janeiro, Brazil}

\acmBooktitle{Proceedings of the 34th ACM International Conference on Multimedia (MM '26), 2026, TODO: location}
\acmISBN{TODO} 

\usepackage{enumitem}
\usepackage{multirow}
\usepackage{booktabs}
\usepackage{array}

\definecolor{Mycolor}{HTML}{0066cc}

\begin{document}

\title{SADL: What to Ignore? A Benchmark for Subject-Aware Distractor Localization}

\author{Cao-Tri Nguyen}
\email{npctri22@apcs.fitus.edu.vn}
\authornote{Equal contribution.}
\affiliation{%
  \institution{University of Science\\Vietnam National University Ho Chi Minh City}
  \city{Ho Chi Minh City}\country{Vietnam}
}

\author{Nguyen-Khoa Luong}
\email{lnkhoa22@apcs.fitus.edu.vn}
\authornotemark[1]
\affiliation{%
  \institution{University of Science\\Vietnam National University Ho Chi Minh City}
  \city{Ho Chi Minh City}\country{Vietnam}
}

\author{Vinh-Tiep Nguyen}
\email{tiepnv@uit.edu.vn}
\affiliation{%
  \institution{University of Information Technology\\Vietnam National University Ho Chi Minh City}
  \city{Ho Chi Minh City}\country{Vietnam}
}

\author{Minh-Triet Tran}
\email{tmtriet@fit.hcmus.edu.vn}
\authornote{Corresponding author.}
\affiliation{%
  \institution{University of Science\\John Von Neumann Institute\\Vietnam National University Ho Chi Minh City}
  \city{Ho Chi Minh City}\country{Vietnam}
}

\renewcommand{\shortauthors}{Nguyen et al.}

\begin{abstract}
Photographs frequently contain \emph{visual distractors} besides foregrounds and backgrounds of the intended subject, competing for attention and weakening composition. While modern editing tools streamline object removal, identifying which objects to remove remains a mostly manual process. Existing saliency models and open-vocabulary detectors operate without subject awareness, failing to adapt to shifting user intent. Furthermore, context-agnostic removal may disrupt the scene's semantic coherence (e.g., keep the person but remove the chair they are sitting on). To address these limitations, we formalize the task of subject-aware distractor localization, which identifies distractors while retaining compositionally essential objects. This paper introduces \textsc{SADL}, the first real-world benchmark for this task, comprising 1,800 subject-aware cases across 1,000 photographs to enable systematic evaluation and facilitate future research. In total, there are 14,617 annotated candidates, including a robust set of 1,938 hard negatives to stress-test exclusion calibration. We evaluate seven proprietary and open-weight Vision-Language Models (VLMs) on a sequential pipeline of distractor classification followed by exclusion filtering, structured around five inclusion factors and three contextual exclusion rules. Our analysis reveals that VLMs are highly capable of identifying distractors, but then over-apply exclusion, which systematically suppresses true distractors at scale. By exposing this critical bottleneck, \textsc{SADL} provides a foundational diagnostic tool to advance subject-conditioned reasoning in multimodal systems. The dataset is publicly released at \url{https://github.com/ct101apcs/SADL}.
\end{abstract}



\begin{teaserfigure}
\centering
\includegraphics[width=\textwidth]{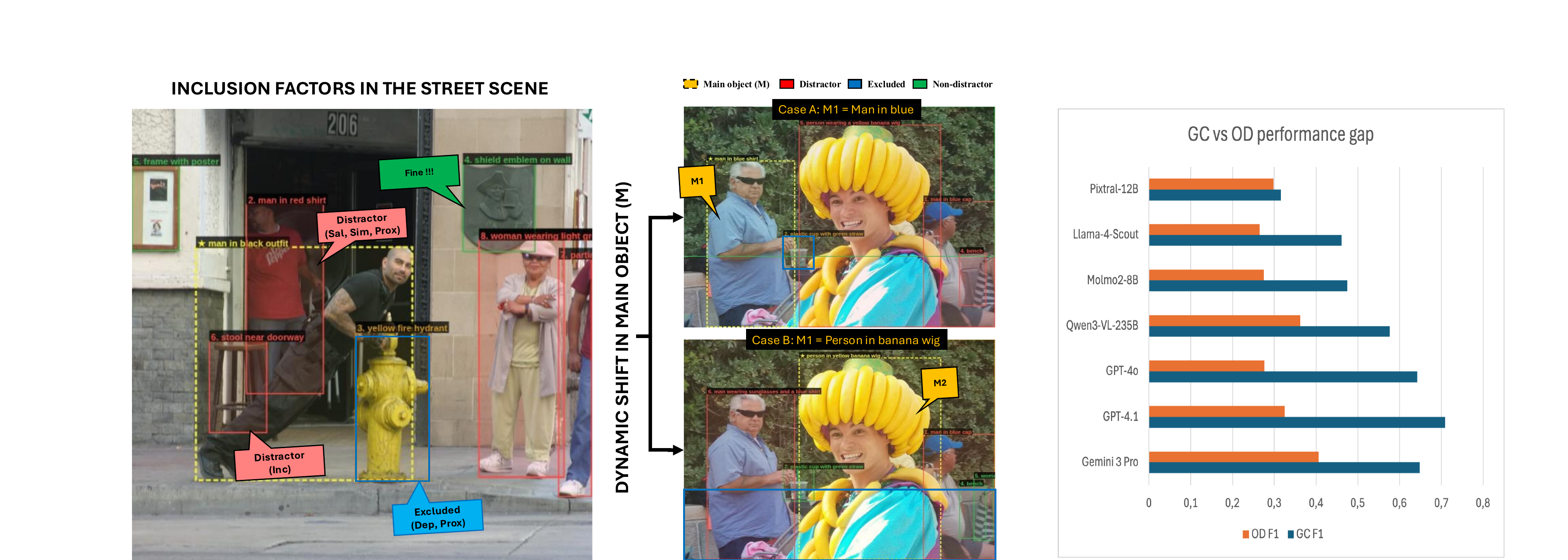}
    \caption{
        \textbf{SADL: Subject-Aware Distractor Localization.}
        \normalfont
        \textit{(Left)} A street scene annotated with all five inclusion factors: candidates are labeled distractor~(D, red), excluded hard negative~(E, blue), or non-distractor~(N, green) with per-factor and per-rule attributions (Sal~$= F_1$, Prox~$= F_2$, Inc~$= F_3$, Dep~$= E_3$). The yellow fire hydrant is excluded ($E_3$, $F_2$): it is spatially prominent but functionally tied to the scene context.
        \textit{(Center)} Subject-relative shift: the same image yields two cases depending on the specified subject $M$. In Case~A ($M{=}$``man in blue'') several surrounding figures and objects compete for attention as distractors; in Case~B ($M{=}$``person in banana wig'') the subject shifts and the distractor set changes accordingly, with some previously neutral candidates now activated. The distractor set is not a property of the image but of the named subject.
        \textit{(Right)} GC vs.\ OD performance gap across all evaluated VLMs. Every model suffers a large drop from GC~F1 to OD~F1, revealing spatial grounding as the primary barrier to end-to-end deployment beyond compositional reasoning ability.
    }
    \Description{Three-panel teaser. Left: annotated street scene showing D/E/N labels with factor attributions. Center: same image with two subject specifications showing how the distractor set shifts. Right: horizontal bar chart comparing GC F1 and OD F1 per VLM.}
    \label{fig:teaser}
    \vspace{1em}
\end{teaserfigure}

\maketitle

\section{Introduction}
\label{sec:intro}

Photographs frequently contain \emph{visual distractors}, objects that compete for attention with the photographer's intended subject or weaken the overall composition. Modern editing tools have made removal trivial once a distractor is identified. However, the preceding step, deciding \emph{which} objects to remove, remains entirely manual. In addition, this process demands compositional expertise that photographers may not have, especially when the scene contains multiple objects with complex relationships.

Existing approaches partially automate detection but share a structural limitation: they produce one fixed distractor suggestion per image with no mechanism for subject-conditioned adaptation. Saliency-based methods~\cite{itti1998saliency,borji2013state} highlight regions with strong visual contrast. Data-driven predictors~\cite{fried2015distractors} learn a generic notion of ``distracting'' from human removal choices. Interactive cleaners~\cite{huynh2023simpson} propagate a user click to visually similar objects. None accepts a specification of the intended main subject, so none can adapt when the user's intent changes: the same object may be a distractor when the woman in the frame is the subject, and an essential compositional anchor when the dog is. Current VLM benchmarks also fail to probe this subject-aware compositional reasoning.

This paper formalizes this problem as \emph{subject-aware distractor localization}. Given an image and a textual description of a main subject $M$, the task is to identify which other objects are distractors \emph{with respect to} $M$. An object is a distractor if it satisfies at least one of five \emph{inclusion factors} (perceptual and relational properties driving attentional competition). It must simultaneously trigger none of the three \emph{exclusion rules}, which enforce contextual retention and override all inclusion factors.

To measure how well current models solve this task, we introduce \textsc{SADL}. This benchmark covers diverse compositional scenarios, including a substantial set of hard negatives (objects that look like distractors but must be retained). We evaluate seven VLMs and four non-VLM baselines under two complementary protocols, where guided classification isolates pure compositional reasoning and open detection measures end-to-end performance under realistic deployment conditions. Evaluation exposes a hierarchy of systematic failure modes, in exclusion calibration, factor detection, and spatial grounding, that binary benchmarks tend to conflate or obscure.

\noindent Our key contributions can be summarized as follows:
\begin{enumerate}[leftmargin=*,nosep]
    \item A formal framework for subject-aware distractor localization with five inclusion factors ($F_1$--$F_5$) and three exclusion rules ($E_1$--$E_3$), grounded in visual attention theory and expert photography, enabling process-level error decomposition by root cause.
    \item \textsc{SADL}, the first real-world benchmark for this task: 1,800 subject-aware cases across 1,000 images, 14,617 annotated candidates, 1,938 hard negatives, and a dual-protocol design isolating compositional reasoning from spatial grounding.
    \item Controlled ablation and IAA-stratified analysis exposing four empirically distinct failure modes (exclusion over-triggering, output contamination, factor detection failure, and spatial grounding failure), each with a distinct diagnostic signature and suggested remediation direction.
\end{enumerate}

\section{Related Work}
\label{sec:related}

\paragraph{Distractor Detection and Scene Editing}
Classical saliency models~\cite{itti1998saliency,itti2001computational,borji2013state} and data-driven predictors~\cite{fried2015distractors,kumain2025saliency_review,zhang2018detecting} produce one fixed distractor map per image with no subject-conditioned adaptation. Editing tools and segmentation models (SAM~\cite{kirillov2023segment}, SAM~2~\cite{ravi2024sam2}, InstructPix2Pix~\cite{brooks2023instructpix2pix}, SimpSON~\cite{huynh2023simpson}) assume a completed upstream identification step. \textsc{SADL} targets exactly this missing step, formalizing subject-aware identification with multi-factor annotation and hard negatives.

\paragraph{Object Grounding and Subject-Aware Reasoning}
Referring expression comprehension~\cite{kazemzadeh2014referitgame,yu2016modeling} localizes a specified entity; our task is the inverse, identifying objects that \emph{compete with} that entity. Open-vocabulary detectors~\cite{liu2023grounding} and grounding models~\cite{xiao2024florence2} lack subject-aware exclusion reasoning; treating all non-subject objects as distractors yields near-random precision. Works that add synthetic distractors to VLM inputs~\cite{bae2025idis,ma2025caution,deng2025words} use distractors as a confound variable, not an identification target; \textsc{SADL} uses real photographs where multi-factor relationships and contextual retention must be jointly reasoned.

\paragraph{Diagnostic Benchmarks for MLLMs}
MathGLANCE~\cite{sun2025mathblind} and UReason~\cite{yang2026ureason} decouple visual perception from reasoning in MLLMs, exposing capability gaps that aggregate metrics obscure. Our dual-protocol design, GC isolating compositional reasoning and OD measuring end-to-end grounding, applies this diagnostic principle to subject-aware distractor localization.

\section{Problem Definition}
\label{sec:problem}

Let $I$ denote an image and $T$ a textual description specifying a main subject $M$. Let $\mathcal{O} = \{o_1, \ldots, o_N\}$ be the set of all detected objects in $I$ excluding $M$. An object $o_i$ is classified as a \textbf{visual distractor} with respect to $M$ if and only if it satisfies at least one inclusion factor and triggers no exclusion rule:
\begin{equation}
  \text{Distractor}(o_i \mid M, I) = \Bigl[\bigvee_{k=1}^{5} F_k(o_i)\Bigr] \wedge \Bigl[\bigwedge_{j=1}^{3} \neg E_j(o_i)\Bigr].
  \label{eq:distractor}
\end{equation}

\begin{table}[t]
\centering
\caption{Comparison of distractor identification approaches. \textsc{SADL} uniquely combines subject awareness, contextual exclusion, and hard negatives.}
\label{tab:comparison}
\resizebox{\linewidth}{!}{
\begin{tabular}{lcccc}
\toprule
Approach & Subject-Aware & Contextual Excl. & Hard Negatives & Input \\
\midrule
Saliency~\cite{itti1998saliency,borji2013state} & No & No & None & Image \\
Fried et al.~\cite{fried2015distractors} & No & No & None & Image \\
SimpSON~\cite{huynh2023simpson} & No & No & None & Image+Click \\
\textbf{Ours (\textsc{SADL})} & Yes & Yes & 1,938 & Image+Text \\
\bottomrule
\end{tabular}
}
\end{table}

Tables~\ref{tab:comparison} and~\ref{tab:factors} define the benchmark's scope. The five inclusion factors, grounded in visual attention theory and expert photography, cover the principal mechanisms by which objects compete for attention relative to $M$. Any triggered exclusion rule ($E_1$--$E_3$) overrides all inclusion factors, retaining objects that extend the subject's identity ($E_1$), constitute expected scene fabric ($E_2$), or are functionally indispensable to the subject's activity ($E_3$).

\begin{table}[t]
\caption{Inclusion factors and exclusion rules. Any triggered $E_j$ overrides all $F_k$.}
\label{tab:factors}
\centering
\resizebox{\linewidth}{!}{%
\begin{tabular}{@{}lp{6.4cm}@{}}
\toprule
\textbf{Factor / Rule} & \textbf{Definition} \\
\midrule
$F_1$ Visual Saliency     & Captures attention via high color contrast, luminance, or texture novelty \\
$F_2$ Spatial Proximity   & Spatially adjacent to $M$; proximity triggers attentional competition (Gestalt proximity) \\
$F_3$ Semantic Incongruity & Out-of-context relative to the scene; expectation violations capture top-down attention \\
$F_4$ Categorical Similarity & Shares the same semantic category as $M$; causes referential ambiguity in VLMs \\
$F_5$ Scale Dominance     & Occupies disproportionately large area relative to $M$, aggressively displacing focus \\
\midrule
$E_1$ Attribute of $M$   & Held by, worn by, or physically attached to $M$ \\
$E_2$ Neutral Environmental & Contextually expected background; too small or numerous to compete individually \\
$E_3$ Functional Dependency & Removing it would make $M$'s depicted activity logically incoherent \\
\bottomrule
\end{tabular}%
}
\end{table}

\paragraph{Factor coverage}
The five factors are jointly sufficient: any real-world distractor falls into at least one category. Marginal rates ($F_1$: 26.1\%, $F_2$: 30.0\%, $F_3$: 13.7\%, $F_4$: 13.8\%, $F_5$: 16.2\%) confirm no single factor dominates; a detailed breakdown of factor combinations is provided in the supplementary material (\texttt{sadl\_supp.pdf}).

\section{\textsc{SADL} Dataset Construction}
\label{sec:benchmark}

\subsection{Image Collection}
\label{sec:collection}

We collected 1,000 high-resolution ($\geq$1024\,px) images from public datasets (COCO~\cite{lin2014microsoft}, Winoground~\cite{thrush2022winoground}, Visual Genome~\cite{krishna2017visualgenome}, Open Images~\cite{kuznetsova2020openimages}), stratified across eight scene categories to ensure diverse compositional structures: street/outdoor (170), indoor (160), nature (140), portrait (150), event (130), sports (100), product/still-life (75), and artistic (75). Images with zero distractors (clean portraits, isolated subjects) are intentionally included to penalize models that over-segment indiscriminately.

\paragraph{Dynamic subject specification}
A single image yields \textbf{multiple main-subject cases}, each defined by a distinct $T$ specifying a different $M$. An image of a woman walking a dog yields separate cases for $M{=}$``the woman'' and $M{=}$``the dog'', with the distractor set shifting accordingly (Figure~\ref{fig:teaser}). These images yield \textbf{1,800 annotated cases} (mean 1.80 cases/image): 30.0\% single-case, 61.8\% two-case, 6.6\% three-case, and 1.6\% four-or-more-case images.

\subsection{Hybrid Annotation Pipeline}
\label{sec:annotation}

\paragraph{Stage 1: Agentic pre-labeling}
Qwen3-VL-32B-Instruct~\cite{qwen2025qwen3} generates candidate object labels per image; SAM~3~\cite{ravi2024sam3} segments each candidate to obtain masks and bounding boxes. The same model proposes 1--5 main-subject cases per image. Qwen3-VL-32B-Instruct is excluded from evaluation to prevent annotation--evaluation circularity; the gold ground truth reflects majority human vote.

\paragraph{Stage 2: Human annotation}
A human reviewer checks all labels for vague or incorrect entries and adjusts bounding boxes where necessary. Two annotators independently label each candidate with all five inclusion factors and three exclusion rules. The final binary distractor label is derived from Eq.~\eqref{eq:distractor}. An adjudicator with professional photography experience resolves all disagreements; the final GT is the majority vote of the three reviewers. Factor-level recall columns in Section~\ref{sec:exp_factors} measure behavioral adherence to the annotation schema; they characterize output patterns and are not claims about latent model representations.

\subsection{Inter-Annotator Agreement}
\label{sec:agreement}

Overall binary distractor agreement between the two primary raters reaches Cohen's $\kappa$~\cite{cohen1960coefficient} $= 0.534$ (raw agreement 77.0\%). Per-factor Krippendorff's $\alpha$~\cite{krippendorff2018content} reveals a clear objectivity spectrum: geometric and categorical factors ($F_2$, $F_4$, $F_5$: $\alpha \geq 0.81$) achieve substantially higher agreement than perceptual factors ($F_1$, $F_3$: $\alpha \approx 0.57$). Both raters agree more strongly with the adjudicated gold standard ($\kappa = 0.674$ and $0.718$) than with each other, confirming adjudication resolves systematic biases. Full IAA tables are provided in the supplementary material (\texttt{sadl\_supp.pdf}).

\subsection{Dataset Statistics}
\label{sec:stats}

\begin{figure*}[t]
  \centering
  \makebox[\linewidth][c]{\includegraphics[width=1.2\linewidth]{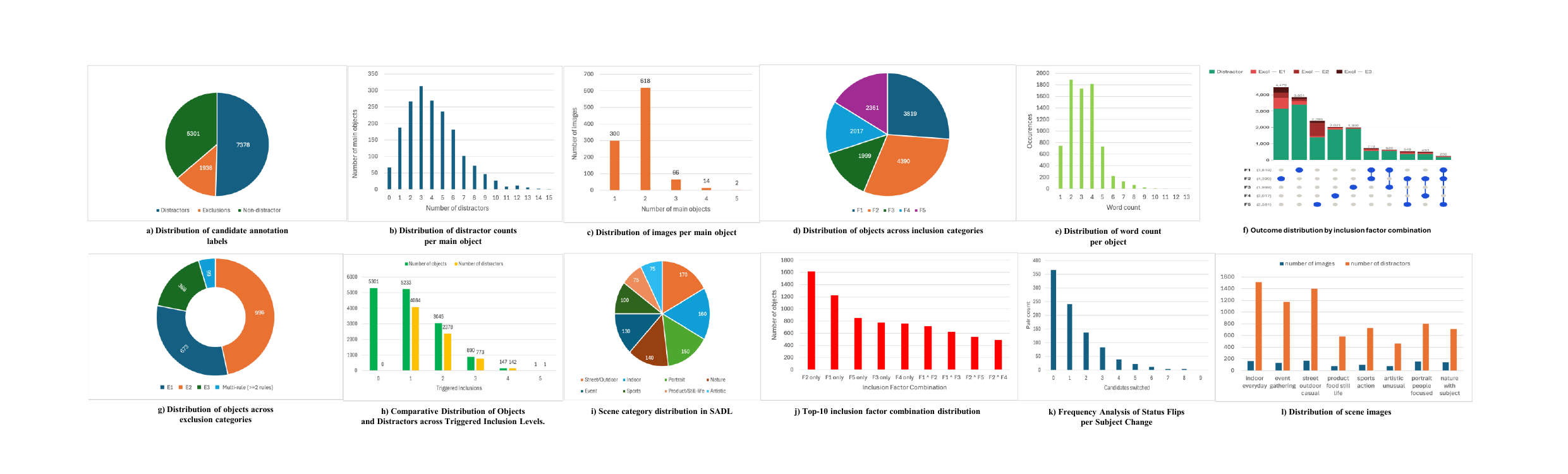}}
  \caption{\textnormal{\textbf{Dataset statistics of \textsc{SADL}.}
  Panels show: (a)~label distribution; (b)~distractors per object; (c)~images per main object; (d)~objects per inclusion factor; (e)~subject description word count; (f)~UpSet plot of factor-combination outcomes; (g)~exclusion-rule distribution; (h)~distractor yield vs.\ triggered inclusion level; (i)~scene category distribution; (j)~top-10 factor combinations; (k)~status-flip frequency per subject change; (l)~scene image distribution.}}
  \label{fig:statistic}
  \vspace{-4mm}
\end{figure*}

Figure~\ref{fig:statistic} provides a comprehensive overview of \textsc{SADL}'s composition and diversity across twelve dimensions. The 14,617 annotated candidates are partitioned into three label classes: 7,378 distractors (D, 50.5\%), 1,938 hard-negative exclusions (E, 13.3\%), and 5,301 non-distractors (N, 36.3\%) (panel~a).

\paragraph{Candidate and factor diversity}
The number of distractors per main object follows a right-skewed distribution peaking at 3--4 (panel~b), while most images contain one or two main objects (panel~c). Coverage is broadly balanced across the five inclusion factors $F_1$--$F_5$ (panel~d). Object descriptions are concise, averaging 4.3 words, with 90\% comprising 1--6 words (panel~e), and model accuracy varies by at most 8 percentage points across length bins, ruling out a systematic length-based difficulty confound. The UpSet plot (panel~f) shows that single-factor activations dominate, with $F_2$ (spatial proximity) and $F_1$ (visual similarity) as the most frequent sole triggers, while multi-factor co-activations are rarer but present throughout.

\paragraph{Exclusion structure and multi-case sensitivity}
Among the 1,938 hard negatives, rules $E_1$ and $E_3$ account for the majority, with fewer than 5\% of objects triggering multiple rules simultaneously (panel~g). Distractor yield grows sub-linearly with the number of satisfied inclusion factors (panel~h), indicating that co-occurring factors are not independently additive. In multi-case images, 59.8\% of case pairs contain at least one candidate that switches distractor status across subjects (panel~k), directly quantifying cross-subject compositional sensitivity.

\paragraph{Scene coverage}
\textsc{SADL} spans eight broad scene categories balanced between indoor and outdoor settings (panels~i,~l). The top-10 factor combinations (panel~j) confirm $F_2$-only as the single most common pattern, with $F_1{\times}F_2$ co-activation as the leading two-factor combination.

\section{Evaluation Framework}
\label{sec:eval}

\subsection{Evaluation Protocols}
\label{sec:protocols}

\paragraph{Guided Classification (GC)}
The model receives the image, description $T$, and a pre-enumerated candidate list $\mathcal{O}$ as text labels. For each candidate it predicts a three-class label (\textit{distractor}~D, \textit{excluded}~E, or \textit{non-distractor}~N) along with per-factor ($F_1$--$F_5$) and per-rule ($E_1$--$E_3$) binary attributions. GC isolates \emph{pure compositional reasoning} with no localization step.

\paragraph{Open Detection (OD)}
The model receives only the image and $T$, freely enumerating distractor candidates as short text labels with optional bounding-box hints. SAM~3~\cite{ravi2024sam3} converts each label into a segmentation mask; Hungarian matching at IoU@0.5 against GT masks determines TP, FP, and FN. OD measures \emph{end-to-end deployment performance}.

The \textbf{GC--OD gap} ($\Delta$F1 = OD F1 $-$ GC F1, negative for all models) decomposes the shortfall into the component attributable to classification (measurable via GC) and the residual attributable to spatial grounding and detection coverage.

\subsection{Metrics and Baselines}
\label{sec:metrics}
\label{sec:baselines}

\paragraph{Metrics}
Each candidate is assigned one of three ground-truth labels: distractor (D), hard-negative exclusion (E), or non-distractor (N). We report per-class recall, \textbf{D-Rec}, \textbf{E-Rec}, and \textbf{N-Rec}, to expose class-level sensitivity. \textbf{Average Recall} $\text{AR} = \tfrac{1}{3}(\text{D-Rec} + \text{E-Rec} + \text{N-Rec})$ is the primary metric; it weights exclusion judgment equally with detection, so a model that ignores exclusion entirely scores at most 33\%. \textbf{DE-GMean} $= \sqrt{\text{D-Rec} \times \text{E-Rec}}$ isolates the D/E trade-off: if either recall collapses to zero, so does the score. \textbf{GC F1} and \textbf{OD F1} are distractor-class F1 scores under each protocol; their difference $\Delta$F1 (always negative) quantifies the spatial grounding bottleneck. Rankings are stable across weighting variants ($\rho \geq 0.93$) and IoU thresholds 0.25--0.75 ($\rho = 1.0$).

\paragraph{Baselines}
Seven VLMs spanning three proprietary models (GPT-4.1~\cite{openai2024gpt41}, Gemini~3~Pro~\cite{team2025gemini3}, GPT-4o~\cite{openai2024gpt4o}) and four open-weight models (Qwen3-VL-235B~\cite{qwen2025qwen3}, Llama-4-Scout~\cite{meta2025llama4}, Molmo2-8B~\cite{deitke2025molmo2}, Pixtral-12B~\cite{mistral2024pixtral}) are evaluated zero-shot under both protocols. The Random assignment baseline, which randomly assigns D/E/N labels based on the dataset distribution and ignores all factors and rules, is included as a lower bound. Four non-reasoning methods (GroundingDINO~\cite{liu2023grounding}, U\textsuperscript{2}-Net~\cite{qin2020u2net}, CLIP-text and CLIP-image~\cite{radford2021learning}) yield binary D/N outputs with no exclusion reasoning and are reported separately in Table~\ref{tab:baselines_nonvlm}. A human reference (180 GC cases, blind to gold labels) provides an upper bound.

\section{Experiments and Analysis}
\label{sec:exp}

\subsection{Main Results}
\label{sec:exp_main}

Tables~\ref{tab:results} and~\ref{tab:baselines_nonvlm} report GC and OD performance; SAM~3 success rate is included per VLM as a proxy for label-to-mask conversion quality.

\begin{table*}[t]
\caption{VLM performance across all 14,617 candidates under GC and OD protocols. \textbf{GC input}: (image, subject description $T$, pre-enumerated candidate list) $\to$ D/E/N label per candidate. \textbf{OD input}: (image, $T$) $\to$ free-text labels, grounded to masks by SAM~3. \textbf{SAM3\%}: fraction of enumerated labels converted to a non-empty mask. \textbf{FN\textsubscript{loc}\%}: OD false negatives from spatial grounding failure. $\boldsymbol{\Delta}$\textbf{F1} = OD F1 $-$ GC F1. Metrics defined in Section~\ref{sec:metrics}.}
\label{tab:results}
\centering
\resizebox{\textwidth}{!}{%
\begin{tabular}{l|cccccc|ccc|c}
\toprule
 & \multicolumn{6}{c|}{\textbf{Guided Classification (GC)}} & \multicolumn{3}{c|}{\textbf{Open Detection (OD, IoU@0.5)}} & \\
\textbf{Model} & \textbf{AR\%}$\uparrow$ & \textbf{DE-GMean}$\uparrow$ & \textbf{GC F1}$\uparrow$ & \textbf{D-Rec\%}$\uparrow$ & \textbf{E-Rec\%}$\uparrow$ & \textbf{N-Rec\%}$\uparrow$ & \textbf{OD F1}$\uparrow$ & \textbf{SAM3\%} & \textbf{FN\textsubscript{loc}\%}$\downarrow$ & $\boldsymbol{\Delta}$\textbf{F1}$\uparrow$ \\
\midrule
Gemini 3 Pro~\cite{team2025gemini3}   & \textbf{63.8} & \textbf{69.3} & 0.648 & 53.5 & \textbf{89.8} & 48.2 & \textbf{0.406} & 82.5 & 82.9 & $-$0.242 \\
GPT-4.1~\cite{openai2024gpt41}        & 62.3 & 63.0 & \textbf{0.709} & \textbf{65.1} & 61.0 & 60.9 & 0.325 & 77.1 & 96.4 & $-$0.384 \\
GPT-4o~\cite{openai2024gpt4o}         & 58.0 & 61.6 & 0.642 & 56.7 & 66.9 & 50.4 & 0.276 & 87.9 & 95.6 & $-$0.366 \\
Qwen3-VL-235B~\cite{qwen2025qwen3}    & 54.0 & 55.4 & 0.576 & 46.3 & 66.4 & 49.3 & 0.362 & 99.5 & \textbf{71.1} & $-$0.214 \\
Molmo2-8B~\cite{deitke2025molmo2}     & 46.9 & 33.4 & 0.475 & 35.5 & 31.4 & \textbf{73.8} & 0.275 & 91.3 & 93.0 & $-$0.200 \\
Llama-4-Scout~\cite{meta2025llama4}   & 43.7 & 39.8 & 0.461 & 34.0 & 46.5 & 50.7 & 0.265 & 99.3 & 85.8 & $-$0.196 \\
Pixtral-12B~\cite{mistral2024pixtral} & 40.7 & 27.4 & 0.316 & 20.0 & 37.4 & 64.6 & 0.298 & 97.8 & 77.4 & \textbf{$-$0.018} \\
\midrule
Random assignment           & 33.0 & 26.5 & 0.500 & 49.9 & 14.0 & 35.2 & -- & -- & -- & -- \\
Human                       & 87.3 & 84.1 & 0.842 & 79.3 & 89.1 & 93.4 & -- & -- & -- & -- \\
\bottomrule
\end{tabular}%
}
\end{table*}

\begin{table}[t]
\caption{Non-reasoning methods under GC. All produce binary D/N outputs (no E class). GC~F1 for GroundingDINO treats all detections as distractors. OD uses each model's native output without SAM~3.}
\label{tab:baselines_nonvlm}
\centering
\resizebox{\linewidth}{!}{%
\begin{tabular}{lcccccc}
\toprule
\textbf{Method}& \textbf{GC F1}$\uparrow$ & \textbf{D-Rec\%}$\uparrow$ & \textbf{N-Rec\%}$\uparrow$ & \textbf{OD F1}$\uparrow$ \\
\midrule
GroundingDINO~\cite{liu2023grounding} & \textbf{0.580} & \textbf{60.8} & \textbf{53.1} & \textbf{0.206} \\
U\textsuperscript{2}-Net~\cite{qin2020u2net}& 0.551 & 56.6 & 49.8 & 0.042 \\
CLIP-text~\cite{radford2021learning}  & 0.480 & 46.3 & 50.7 & -- \\
CLIP-image~\cite{radford2021learning} & 0.471 & 45.5 & 47.8 & -- \\
\bottomrule
\end{tabular}%
}
\end{table}

\paragraph{Overall performance}
On AR and DE-GMean in GC settings, proprietary VLMs substantially outperform Random assignment, but still witness a large gap to human performance. This gap is even larger under OD, which is the nearest to real-world deployment conditions. Non-reasoning methods seem to score competitively with VLMs on GC F1, but they are not directly comparable due to the difference in class structure (binary D/N vs.\ three-class D/E/N). The fairer comparison is to look at Ablation D-Rec in Table~\ref{tab:ablation}, where non-reasoning methods score at most 60.8\% recall, while VLMs reach up to 88.9\% when exclusion rules are removed.

\paragraph{Spatial grounding bottleneck}
GC~F1 and OD~F1 rankings diverge across models: Qwen3-VL-235B ranks 4th by GC~F1 but rises to 2nd by OD~F1, while GPT-4.1 holds the highest GC~F1 yet falls to 3rd under OD. The divergence is driven by localization rather than reasoning quality: models producing concise, noun-phrase outputs ground more successfully via SAM~3 and lose less in OD than those with verbose or imprecise descriptions. Most of the F1 shortfall is attributable to label-to-mask conversion failure rather than SAM~3 limitations.

\subsection{Exclusion Over-Triggering}
\label{sec:exp_ablation}

\begin{table}[t]
\caption{Exclusion criterion ablation. \textbf{GC D-Rec}: distractor recall under the full prompt. \textbf{Ablation D-Rec}: recall when $E_1$--$E_3$ are removed. \textbf{Redistrib.\%}: of GT-distractors wrongly labeled \textit{excluded} under the GC prompt, the fraction that correctly flips to \textit{distractor} when exclusion rules are removed. High redistribution ($>$90\%) confirms E over-triggering as the causal mechanism, not detection failure.}
\label{tab:ablation}
\centering
\resizebox{\linewidth}{!}{%
\begin{tabular}{lcccc}
\toprule
\textbf{Model} & \textbf{GC D-Rec\%}$\uparrow$ & \textbf{Ablation D-Rec\%}$\uparrow$ & \textbf{$\Delta$D (pp)}$\uparrow$ & \textbf{Redistrib.\%}$\uparrow$ \\
\midrule
GPT-4.1        & 65.1 & 88.9 & $+$23.8 & 95.2 \\
Gemini 3 Pro   & 53.5 & 86.7 & $+$33.2 & 90.7 \\
GPT-4o         & 56.7 & 89.7 & $+$33.0 & 95.9 \\
Qwen3-VL-235B  & 46.3 & 92.2 & $+$45.9 & 96.8 \\
Llama-4-Scout  & 34.0 & 77.8 & $+$43.8 & 78.0 \\
Pixtral-12B    & 20.0 & 32.2 & $+$12.2 & 33.1 \\
\bottomrule
\end{tabular}%
}
\end{table}

\paragraph{E over-triggering}
Removing exclusion rules from the prompt recovers nearly all distractors that frontier models previously suppressed, establishing over-triggering as the root cause of their D-Rec deficit (Table~\ref{tab:ablation}). Models apply exclusion criteria at scene-level plausibility rather than anchoring them to the named subject. Binary benchmarks cannot surface this failure mode, as excluded candidates are indistinguishable from true negatives without the three-class distinction.

\paragraph{Factor detection failure}
Pixtral's errors persist even after exclusion rules are removed, pointing to a different root cause. Many true distractors receive no triggered inclusion factors, meaning the model fails at the perceptual grounding step rather than the exclusion step. Weaker models need factor-level improvements; exclusion recalibration alone is insufficient. Llama's cross-protocol results further show that E over-triggering is a consistent failure across both GC and OD.

\subsection{Process-Level Diagnostics}
\label{sec:exp_factors}

\begin{table}[t]
\caption{Factor-level recall under the ablation prompt (GC mode). \textbf{MF-Rec}: mean recall across $F_1$--$F_5$ on GT-positive candidates. Columns measure each model's ability to identify each type of attentional competition. $F_1$ = visual saliency, $F_2$ = spatial proximity, $F_3$ = semantic incongruity, $F_4$ = categorical similarity, $F_5$ = scale dominance.}
\label{tab:factor_recall}
\centering
\resizebox{\linewidth}{!}{%
\begin{tabular}{lcccccc}
\toprule
\textbf{Model} & \textbf{MF-Rec\%}$\uparrow$ & \textbf{F1\%}$\uparrow$ & \textbf{F2\%}$\uparrow$ & \textbf{F3\%}$\uparrow$ & \textbf{F4\%}$\uparrow$ & \textbf{F5\%}$\uparrow$ \\
\midrule
Gemini 3 Pro   & \textbf{73.2} & \textbf{84.9} & 83.7 & 28.1 & \textbf{94.8} & \textbf{74.5} \\
GPT-4.1        & 65.4 & 84.0 & 81.0 & 42.9 & 69.3 & 49.6 \\
GPT-4o         & 63.1 & 59.6 & 76.8 & \textbf{58.9} & 65.1 & 55.2 \\
Qwen3-VL-235B  & 60.4 & 84.1 & \textbf{91.5} & 28.6 & 64.7 & 33.2 \\
Llama-4-Scout  & 48.2 & 39.4 & 76.6 & 53.1 & 55.9 & 16.1 \\
Molmo2-8B      & 32.4 & 52.7 & 46.1 & 20.8 & 36.8 &  5.4 \\
Pixtral-12B    & 31.2 & 30.4 & 65.5 &  2.8 & 49.4 &  8.0 \\
\midrule
Human  & 90.1 & 90.4 & 91.2 & 82.1 & 92.6 & 94.4  \\
\bottomrule
\end{tabular}%
}
\end{table}

\paragraph{Factor difficulty is non-uniform and model-stable}
Categorical similarity ($F_4$) is consistently the easiest factor, likely because category matching aligns with VLM pre-training. Semantic incongruity ($F_3$) is the hardest, with systematic under-detection across all models. Spatial proximity ($F_2$) produces the most false positives, as models over-project proximity beyond true distractor candidates, particularly in the common $F_1 \wedge F_2$ co-occurrence.

\paragraph{Models fail where annotators agree most}
IAA-stratified analysis inverts the usual assumption: model errors concentrate on \emph{geometric} candidates ($F_2$, $F_4$, $F_5$), where human annotators agree most strongly, rather than on perceptually ambiguous ones. This points to genuine capability gaps in geometric relational reasoning, not annotation noise, and highlights a direction where current VLMs fall furthest behind human performance.

\subsection{Failure Mode Taxonomy}
\label{sec:failure_modes}

Four empirically distinct failure modes are identified across models (non-mutually exclusive). For each, we note the diagnostic signature and a suggested remediation direction; these directions are hypotheses motivated by the observed pattern, not validated interventions.

\paragraph{$E$ over-triggering (GPT-4.1, Gemini~3~Pro, GPT-4o, Qwen3-VL-235B)}
Redistribution rates of 91--97\% confirm that frontier models systematically misapply exclusion rules at scene-level plausibility rather than anchoring them to the named subject. The most promising remediation direction is decoupling factor detection from exclusion judgment in the prompt structure, and explicitly conditioning each exclusion criterion on the named subject $M$ rather than scene context.

\paragraph{Output contamination (Llama-4-Scout, Molmo2-8B, Pixtral-12B)}
Open-weight models produce structurally inconsistent outputs where exclusion criteria fire without any triggered inclusion factor, a logical precondition violation that silently suppresses true distractors (contamination rate 55--63\%). Correcting these inconsistencies substantially recovers recall for Llama, while frontier models are nearly unaffected. Schema enforcement via structured output formats and adding a factor precondition check to the prompt are candidate directions for future work.

\paragraph{Factor detection failure (Pixtral-12B, Molmo2-8B)}
Pixtral's errors persist even after exclusion rules are removed, with $\geq$54.6\% of true distractors receiving zero triggered inclusion factors. The failure is at the perceptual grounding step rather than the exclusion step, suggesting that factor-targeted fine-tuning or chain-of-thought grounding prompts may be more effective than exclusion recalibration alone.

\paragraph{Spatial grounding failure (all models; worst GPT-4.1, $\text{FN}_\text{loc} = 96.4\%$)}
All VLMs suffer large GC-to-OD drops driven by label-to-mask conversion failure rather than SAM~3 limitations. Models producing concise noun-phrase labels ground more successfully; verbose or ambiguous outputs fail at the grounding step. Adopting noun-phrase output constraints and iterative label-to-mask verification are suggested as practical remediation directions.

Qualitative examples illustrating these failure modes in detail are provided in the supplementary material.

\section{Discussion}
\label{sec:discussion}

The dominant finding, exclusion over-triggering as the causal bottleneck, would be conflated with detection failure under a binary benchmark, where excluded candidates collapse into the negative class. The three-class design is what makes this failure mode observable.

\paragraph{Practical implications}
The benchmark provides process-level attribution of \emph{why} current models fail (E-calibration for frontier models, factor detection for open-weight models, and spatial grounding for all), and the rich candidate set with full factor-level attributions supports fine-tuning on structured distractor reasoning. The factor-and-rule schema functions as a subject-aware attribute decomposition~\cite{an2024perceptionclip}: a model that fully understands the five factors can adapt as user intent changes, for instance shifting from ``remove all audiences'' to ``remove only audiences near the subject,'' which requires proximity reasoning ($F_2$) conditioned on a new $M$. Dynamic subject specification supports this directly by making every annotated image reusable across different subject specifications without re-annotation.

\paragraph{Dataset release and reproducibility}
\textsc{SADL} is publicly released at the project page under CC~BY~4.0 for all annotations; source images retain their original licenses (COCO, Winoground, Visual Genome, Open Images), all of which permit research use. The release includes per-candidate factor/rule labels, bounding boxes, segmentation masks, subject descriptions, evaluation code, and prompt templates. All VLMs are evaluated zero-shot at fixed model snapshots with documented random seeds. Source images are drawn from publicly available datasets; no personally identifiable data was collected, and all annotation work was performed by compensated, consenting annotators.

\paragraph{Limitations}
The 1,000-image scale reflects a deliberate quality-first design; diagnostic power derives from annotation density and dual-protocol evaluation rather than image count, and no evaluated model approaches saturation. The five factors were selected through expert consultation and literature review to cover the principal mechanisms by which objects compete for attention, and their joint coverage is validated by distractor-type saturation across the dataset; the schema can be extended by adding new factors, though we believe the current set already covers the large majority of real-world distracting patterns. The candidate universe was seeded by an agentic pre-labeling step; annotators added only 1.46\% of candidates not proposed by the model and removed 4.04\% as false proposals, confirming near-exhaustiveness. OD detections outside GT are counted as false positives; a small fraction could correspond to valid distractors missed at the proposal stage.

\section{Conclusion}
\label{sec:conclusion}

We introduced \textsc{SADL}, the first diagnostic benchmark for subject-aware visual distractor localization. Controlled ablation reveals that frontier VLMs fail primarily at exclusion calibration, misapplying rules at scene-level plausibility rather than anchoring them to the named subject. Beyond reasoning, spatial grounding forms the primary barrier to end-to-end deployment. Open-weight models additionally suffer from output contamination and factor detection failure as distinct, diagnosable pathologies. \textsc{SADL} is publicly released with annotations, evaluation code, and prompt templates, providing a reproducible process-level diagnostic foundation for subject-aware compositional understanding.

\bibliographystyle{ACM-Reference-Format}
\bibliography{refs}


\end{document}